\documentclass[runningheads]{llncs}
\usepackage[T1]{fontenc}
\usepackage{cite}
\usepackage{amsmath,amssymb,amsfonts}
\usepackage{graphicx}
\usepackage{textcomp}
\usepackage{xcolor}
\usepackage[hyphens]{url}
\def\BibTeX{{\rm B\kern-.05em{\sc i\kern-.025em b}\kern-.08em
    T\kern-.1667em\lower.7ex\hbox{E}\kern-.125emX}}
\usepackage{algorithm}
\usepackage{algpseudocode}
\definecolor{RoyalBlue}{RGB}{65,105,225}
\definecolor{ForestGreen}{RGB}{34,139,34}
\definecolor{BurntOrange}{RGB}{204,85,0}
\usepackage{booktabs}
\usepackage{makecell}
\usepackage{relsize}
\usepackage{float}

\usepackage{newfloat}
\usepackage{listings}
\floatstyle{ruled}
\newfloat{listing}{tb}{lst}{}
\floatname{listing}{Listing}

\def\|#1|{\mathid{#1}}
\newcommand{\mathid}[1]{\ensuremath{\mathit{#1}}}
\def\<#1>{\codeid{#1}}
\protected\def\codeid#1{\ifmmode{\mbox{\sf{#1}}}\else{\sf #1}\fi}
\usepackage{color}
\usepackage{cleveref}
\usepackage{xspace}

\newcommand{\todo}[1]{{\color{red}\bfseries [[#1]]}}
\ifdefined\notodocomments
  \renewcommand{\todo}[1]{}
\fi

\newcommand{\numJavaTrainConflicts}{3424\xspace}
\newcommand{\numJavaTrainRepositories}{67\xspace}
\newcommand{\numJavaTestConflicts}{806\xspace}

\newcommand{\numTotalConflicts}{7938\xspace}
\newcommand{\numTotalRepositories}{1439\xspace}
\newcommand{\numLanguages}{11\xspace}

\newcommand{\numCommercialLlmsMinusOne}{3\xspace}

\newcommand{\numContextLines}{20\xspace}
\newcommand{\modelParameters}{14B\xspace}
\newcommand{\maxInputTokenLimit}{512\xspace}
\newcommand{\maxOutputTokenLimit}{2048\xspace}
\newcommand{\quantizationBits}{4\xspace}
\newcommand{\loraRank}{128\xspace}
\newcommand{\modelTemperature}{0.9\xspace}
\newcommand{\numGenerations}{16\xspace}
\newcommand{\batchSize}{1\xspace}
\newcommand{\gradAccumSteps}{4\xspace}
\newcommand{\trainingDays}{4\xspace}

\newcommand{\formatStrugglingInvalidMin}{5.5\xspace}
\newcommand{\formatStrugglingInvalidMax}{9.3\xspace}
\newcommand{\conservativeResolversMin}{40.1\xspace}
\newcommand{\conservativeResolversMax}{86.1\xspace}

\newcommand{\precaptionspace}{\vspace{-5pt}}
\newcommand{\postgraphicsspace}{\vspace{-5pt}}

\title{Merge-Bench: Resolve Merge Conflicts with Large Language Models}

\author{
\IEEEauthorblockN{}
\IEEEauthorblockA{\textit{Amazon}\\
scheschb@amazon.co.uk}
\and
\IEEEauthorblockN{}
\IEEEauthorblockA{\textit{University of Washington}\\
mernst@cs.washington.edu}
}

\author{Benedikt Schesch\inst{1}\orcidID{0009-0002-2885-3067}\thanks{Work done independently of the author's role at Amazon.} \and
Michael D. Ernst\inst{2}\orcidID{0000-0001-9379-277X}}
\authorrunning{B. Schesch et al.}
\institute{Amazon
\email{scheschb@amazon.co.uk}
  \and
  University of Washington, USA
\email{mernst@cs.washington.edu}}

\begin{document}

\maketitle

\begin{abstract}
This paper applies machine learning to the difficult and important task of
version control merging.
(1) We constructed a dataset, Merge-Bench, of \numTotalConflicts real-world merge
conflict hunks from \numTotalRepositories GitHub repositories.  The ground truth is the merge
resolution that developers committed to the repository.  Our dataset
construction methodology is scalable to arbitrary amounts of data since
no manual labeling is required.
(2) We trained a model, LLMergeJ, to resolve merge conflicts in Java programs. Our approach uses Group Relative Policy Optimization (GRPO), an online reinforcement learning method, to train a Large Language Model (LLM\@).
(3) We performed two evaluations of the performance of LLMs on resolving merge conflicts.
On Java programs, LLMergeJ with
\modelParameters parameters outperforms \numCommercialLlmsMinusOne
commercial LLMs, trailing only Gemini 2.5 Pro.
Across \numLanguages programming languages, commercial LLM performance is
largely stable from language to language.  The best models correctly resolve less
than 60\% of merge conflicts.

\end{abstract}

\section{Introduction}

A merge conflict occurs when a version control system cannot automatically
integrate concurrent changes from different developers. A merge conflict
requires manual resolution, consuming significant developer time and often
causing software defects \cite{BrindescuAJS2020}. Traditional automated
merge tools rely on syntactic and structural heuristics, but they struggle
with semantic conflicts that require understanding code intent and
behavior.
Large Language Models (LLMs) offer a promising approach to automate merge
conflict resolution by leveraging their code understanding capabilities
\cite{SvyatkovskiyFGMDBJSL2022,DinellaMSBNL2023}.

Training and evaluating a merge tool requires a dataset of problems and ground-truth solutions.
We created a new dataset, \textbf{Merge-Bench}, containing
\numTotalConflicts real-world merge conflict hunks from \numTotalRepositories
public GitHub repositories.  The ground truth is the programmer's merge resolution
committed to the repository.
Our code and data are publicly available.\footnote{%
Dataset construction: \url{https://github.com/benedikt-schesch/Merge-Bench-Builder},
Model training: \url{https://github.com/benedikt-schesch/LLMerge},
Evaluation: \url{https://github.com/benedikt-schesch/Merge-Bench}}

Merge-Bench does not suffer from common,
serious problems in other benchmarks (further discussed in the related
work section).
It is \textbf{realistic} because the merge conflicts come from real-world,
widely used repositories.
It suffers little \textbf{data leakage} because the textual conflicts used in its
problem statements do not appear explicitly in the GitHub repository.
(However, the merge resolutions (ground truth) and surrounding context
lines are almost certainly in commercial LLM training data.)
It is \textbf{scalable} because neither dataset creation nor merge tool
evaluation requires human effort, such as creating ground truth.
It is not prone to \textbf{reward hacking}, where the model optimizes the
reward function to the detriment of real-world usefulness.  Reward hacking
is a significant problem with evaluations that run a program's existing
test suite.  We introduce a test-free paradigm for evaluating merge
conflict resolution, along with a normalization procedure for the output of
merging tools.

We developed \textbf{LLMergeJ}, a \modelParameters parameter model trained with Group
Relative Policy Optimization (GRPO)~\cite{ShaoWZXSBZZLWG2025} to resolve merge conflicts in
Java.  LLMergeJ demonstrates the use of Merge-Bench, illustrates the
weaknesses of commercial LLMs on the task of merging, and shows the strength
of reinforcement learning.

Our contributions include:
\begin{itemize}
\item \textbf{Test-free evaluation paradigm for evaluating merges}: A methodology
  that bypasses test execution, enabling scalable training on
  millions of resolved conflicts without reward-hacking vulnerabilities.

\item \textbf{Human-filter-free SWE-RL task}: A
  software-en\-gin\-eer\-ing-oriented reinforcement learning task that combines
  internet-scale data
  with zero human filtering or manual adjustments, enabling truly scalable
  training.

\item \textbf{Merge-Bench}: A scalable benchmark resistant to reward
  hacking and requiring no manual annotation.

\item \textbf{LLMergeJ}: The first model trained with online RL for
  merge conflict resolution, achieving 49\% exact match accuracy and 59\%
  source code match accuracy on real-world conflicts.

\item \textbf{Evaluation of LLMergeJ}:
  Experiments showing that our \modelParameters parameter
  model achieves competitive or better performance than \numCommercialLlmsMinusOne
  commercial LLMs.

\item \textbf{Cross-language evaluation}:
  Experiments showing the performance of 6 state-of-the-art commercial
  models across \numLanguages programming languages.
\end{itemize}

\section{Related Work}

\subsection{Software engineering task benchmarks}

Previous LLM evaluations for code rely on two primary approaches: real code
or interview questions.  Each has fundamental limitations.

\textbf{Real programs.} SWE-Bench~\cite{JimenezYWYPPN2024} validates model outputs by executing project test suites. This approach suffers from a critical vulnerability{\,---\,}models learn to satisfy tests rather than produce correct code. This reward-hacking problem is not hypothetical. GenProg \cite{WeimerNLGF2009} demonstrated that evolutionary algorithms could generate program repairs (patches) satisfying test suites. However, these patches \emph{usually} made the tests pass by deleting functionality or tests or by overfitting to specific inputs \cite{QiLAR2015}. LLMs, too, may ``solve'' programming tasks by exploiting memory vulnerabilities to avoid correctness checks \cite{SakanaAI2025}.

SWE-Bench has similar problems.  Later
researchers~\cite{openai2024swebenchverified} found that SWE-Bench
systematically underestimated AI model capabilities due to faults in
SWE-Bench itself: unreliable development environments causing false test
failures, problem descriptions too vague for AI systems to understand
properly, and overly restrictive unit tests requiring exact matches for
undocumented implementation details. To address these concerns, SWE-Bench
Verified~\cite{openai2024swebenchverified} employed 93 professional
software developers for manual annotation and developed a Docker-based
evaluation harness, reducing the dataset by 78\% from 2,294 to
500 verified samples. This costly process is not scalable.

As a point of comparison, Merge-Bench's \numTotalRepositories repositories
and \numTotalConflicts samples is much larger than SWE-Bench's 12
repositories and 2294 or 500 samples. Merge-Bench is also easy to expand
in size, with no human judgment required.

\textbf{Interview questions.}
LiveCodeBench~\cite{JainHGLYZWSLSS2024} consists of algorithmic problems
from interview platforms like LeetCode. A positive is that these
benchmarks avoid some reward-hacking issues through their simplicity and
robust test suites.  However, they are fundamentally unrealistic and do not represent
the complexity of real repositories where developers must navigate existing
code, understand context, and maintain compatibility with surrounding
systems.

\subsection{Merging benchmarks}

ConflictBench~\cite{ShenM2024} contains 180 hunks, which the authors call
``merging scenarios''.  ConflictBench contains only
hunks where each version has no more than 20 lines of text.
(Our benchmark Merge-Bench
has the same size restriction.)

GitGoodBench~\cite{LindenbauerBZ2025} contains about 337 merge resolution
problems.  The authors'
filtering discarded some merges, including those with more than 8 hunks,
those with conflicts in non-code files, etc.  ConGra~\cite{ZhangSYQ2024}
mined 44,948 conflicts from 34 projects across 4 programming
languages. While ConGra is larger than Merge-Bench in raw conflict count,
it has less diversity in terms of repositories and programming languages.
The dataset's size makes evaluation on commercial LLMs prohibitively
expensive (at least tens of thousands of dollars).
Consequently, ConGra's evaluation focused primarily on smaller open-source
models, with Llama3 8B performing best. Its evaluation protocol
instructs LLMs to always attempt resolution rather than preserving conflicts
when developer intent is ambiguous. Furthermore, ConGra considers a resolution
correct if any similarity metric exceeds 80\%, which does
not guarantee semantic equivalence and may accept functionally incorrect solutions.

\subsection{LLM-based merging tools}

MergeBERT~\cite{SvyatkovskiyFGMDBJSL2022} converts the merge problem into a
classification problem.  It neurally classifies a merge into one of 9 resolution
patterns, such ``choose the left text'', ``choose the right'', ``choose the base'',
``concatenate the left and right'', etc.  The authors claim 63--68\% accuracy.
MergeBERT is not publicly available.
DeepMerge~\cite{DinellaMSBNL2023} outputs a proposed merge, where each line of
the merge is a line from one of the two versions being merged.  DeepMerge's
precision is less than that of jsFSTMerge~\cite{TrindadeTavaresBCS2019}, a
semi-structured merge tool that uses source text and ASTs.  DeepMerge is not
publicly available.

MergeGen~\cite{DongZSLH2023} treats the merge problem as a generation problem rather than a
classification problem (as MergeBERT and DeepMerge did).
The MergeGen authors trained an encoder and decoder and obtained results
slightly better than those of MergeBERT\@.

Gmerge~\cite{ZhangMKPL2022} is a wrapper around GPT-3, where the prompt
engineering embodies $k$-shot learning by incorporating several examples in
the prompt.  Gmerge is not publicly available.
ChatMerge~\cite{ShenYPZ2023} trains a classifier; if the
classifier is unsuccessful, then ChatMerge calls out to ChatGPT to get an answer.

\section{Merge-Bench}
\label{sec:merge-bench}

\subsection{Structure of the benchmark}

% We start with terminology.
In a version control system, a \emph{merge} integrates
two versions of text, where each version was created by independent,
concurrent work.  The two versions of text are called \emph{left} and
\emph{right}; \emph{base} is their common ancestor, which both versions edited.  A
version control system can automatically integrate non-overlapping edits;
however, if both versions edit the same part of the base document, then the
version control system presents merge \emph{conflicts} to the user.
A conflict consists of a set
of \emph{hunks}, each of which represents a minimal part of a document that both
versions edited differently (plus a few nearby context lines that are
identical in base, left, and right).  In each hunk, the left and right text differ.
The textual representation of a hunk is a \emph{3-way diff} that includes
the left, base, and right text separated by conflict markers such as
{\smaller\texttt{<\relax<\relax<\relax<\relax<\relax<\relax<}}, {\smaller\texttt{=======}}, and
{\smaller\texttt{>\relax>\relax>\relax>\relax>\relax>\relax>}}.

The \textbf{Merge-Bench benchmark} consists of a set of merge conflict hunks.
The merge
conflict hunk contains text for left, right, and base.  It also contains pre-context and
post-context, the common text before and after the hunk.  Context is important because resolving a conflict typically
requires examining the surrounding code.  Merge-Bench represents a conflict
exactly as git and diff3 do:  as a 3-way diff with context lines before and after
it.

For each merge conflict hunk, Merge-Bench also contains the ground truth:  the
text that the programmer manually committed to resolve the conflict.
% typically integrating the changes made by left and right.

\subsection{Dataset construction}
\label{sec:dataset-construction}

Public version control repositories are a rich source of merge conflicts.
A version control repository contains a history of merge conflicts and how
programmers resolved them.
% GitHub contains over 420 million repositories, so it is an extensive source
% of merges.

\textbf{Candidate repository selection.} We began with a large pool of
candidate repositories to ensure sufficient diversity and coverage across
programming languages. For most languages, we used the Reaper dataset
\cite{MunaiahKCN2017}, which contains high-quality projects from the
GHTorrent dataset \cite{Gousios2013}. Go, JavaScript,
TypeScript, and Rust are not present in the Reaper dataset; for them we used
the top 1000 most-starred repositories from GitHub for each language. For
Java specifically, we reserved the first 1000 most-starred repositories as
candidates for training; the following 200 repositories are candidates.

\textbf{Merge collection.} From each candidate repository, we
systematically collected merge conflicts by analyzing up to 1000 branches
per repository, including the main branch, feature branches, and deleted
branches (which remain accessible through GitHub's API\@). Merges from
non-main branches are particularly valuable because they tend to be more
challenging than merges on the main branch \cite{ScheschFYRE2024}. We
replayed each merge using git to obtain conflicts. We sampled hunks from
each repository
% 10-100
to obtain 600--800 hunks per language. We kept the number of sampled merges
per repository low to increase dataset diversity.

Many candidate repositories did not contribute to the final dataset due to two primary factors: (1) repositories that became inaccessible or were deleted after our initial selection (due to its age, Reaper now contains many deleted repositories), and (2) repositories that contained no merge conflicts in our sampled branches or whose conflicts were filtered out during filtering.

\textbf{Resolution extraction and filtering.}
Each hunk in Merge-Bench is computed using a context size of 20 lines.

Merge-Bench is a set of conflicting hunks, each of which a merge tool
analyzes independently. We obtained the merge resolution of each hunk
by setting the context size up to \numContextLines lines before and after
each conflict, but we prevented it from spanning multiple conflicts.
To prevent misidentifying
the resolution from the merged code, we discarded hunks
with missing or repeated context lines in the ground truth.  We also
discarded hunks whose resolution is larger than the sum of left, base,
and right, since such hunks likely contain new code beyond what was in the
files being merged.
Due to our limited computational resources, we enforced size constraints:
we removed hunks where the left, right, base, or ground truth text exceeds
\numContextLines lines, or where the merge conflict itself exceeds
\maxInputTokenLimit tokens (using the DeepSeek R1 tokenizer).
This conservative filtering ensures high-quality data while maintaining full automation, requiring no manual verification or annotation.

% Finally, we performed a coherence check: resolutions that result in hunks more than two lines longer than the original full conflict block are very likely to involve refactoring during the resolution and thus filtered out since they are unpredictable due to the many ways there are to write the same code.
% \mde{Avoid the ambiguous term ``conflict''.  Be specific:  if you mean
%   hunk, say hunk.}

The algorithms that build Merge-Bench can be run on more repositories and with different parameters, including providing complete files or complete repositories to the LLM.

%% \begin{figure}
%% \centering
%% \setlength{\tabcolsep}{.5em}
%% \begin{tabular}{l|r|r}
%% \textbf{Language} & \textbf{\# Conflict hunks} & \textbf{Repos} \\
%% \hline
%% C & 630 & 62 \\
%% C++ & 787 & 150 \\
%% C\# & 777 & 172 \\
%% Go & 676 & 87 \\
%% Java & 806 & 44 \\
%% JavaScript & 759 & 157 \\
%% PHP & 574 & 109 \\
%% Python & 761 & 184 \\
%% Ruby & 653 & 157 \\
%% Rust & 716 & 147 \\
%% TypeScript & 799 & 170 \\ \hline
%% Total & 7938 & 1439
%% \end{tabular}
%% \precaptionspace
%% \caption{Size of the Merge-Bench dataset, by language.}
%% \label{tab:language_stats}
%% \end{figure}

% This is uglier but saves many lines.
\begin{figure}[t]
\setlength{\tabcolsep}{2pt}
  \begin{tabular}{l|cccccccccccc}
\textbf{Language} &
C &
C++ &
C\# &
Go &
Java &
\makecell{Java\\Script} &
PHP &
Python &
Ruby &
Rust &
\makecell{Type\\Script} &
Total \\ \hline

\textbf{\#
% conflict
hunks} &
630 &
787 &
777 &
676 &
806 &
759 &
574 &
761 &
653 &
716 &
799 &
7938
\\

\textbf{\#
% repositories
repos}
&
62 &
150 &
172 &
87 &
44 &
157 &
109 &
184 &
157 &
147 &
170 &
1439

  \end{tabular}
  
\precaptionspace
\caption{Size of the Merge-Bench dataset, by language.}
\label{tab:language_stats}
\end{figure}

\textbf{Final dataset composition.} The final dataset (\cref{tab:language_stats}) contains \numTotalConflicts hunks across \numLanguages programming languages, with an average of 5.5 hunks per repository.

\section{Code Comparison Methods}

When a tool produces code, the generated code must be compared against the
ground truth code.  Here are ways to do the
comparison.

\textbf{Textual.}  The generated code is considered correct if it
exactly matches the ground truth.  This approach is scalable and
language-agnostic.  However, it is overly restrictive and underestimates
the quality of the generated code.  It rejects semantically equivalent code
that differs only in formatting, comments, etc.

\textbf{Source code.}
The generated code and ground truth are normalized by comment removal and
standardized formatting.  This is similar to comparing ASTs (abstract
syntax trees or parse trees).
Requiring only a parser per programming language,
this approach is scalable and less restrictive than textual comparison.
However, it still rejects some semantically equivalent code that uses
different variable names or is structurally different.
In other words, it is a proxy for, and lower bound on, correctness.

% \todo{What about AST comparison, or an approximation of AST comparison?
% Goal: determine whether the difference is only a renaming.
% \begin{enumerate}
% \item After whitespace normalization:
% \item Do a character-granularity diff.
% \item Collect all the difference hunks.
% \item If any hunk is not identifier-to-identifier, give up: the ASTs are not equivalent.
% \item Determine an A-to-B mapping that unions all the identifier diffs.
% \item If that mapping contains internal inconsistencies (foo-to-bar and
% foo-to-baz, or foo-to-quux and bar-to-quux), give up: it's not a renaming.
% \item Apply the mapping to A.
% \item Compare the result to B.  If same, it's a renaming and the ASTs are equivalent.
% \end{enumerate}
% }

\textbf{Semantic.}  The program equivalence problem (determining whether two programs have identical behavior) is undecidable \cite{Rice1953}.
Approximating it requires sophisticated, non-scalable program verification.
It requires access to all libraries
used in the program, which can be difficult for real-world software.
% All of these problems relate to evaluating implementation correctness,
% which has an unambiguous answer; by contrast, code quality and style are
% subjective, so we do not consider it in this paper.

\textbf{Testing.}
A test suite can determine that two programs behave the same on a specific
finite set of outputs, according to a finite number of assertions.
% A positive aspect of testing is that it
Testing
accepts semantically equivalent
programs that are syntactically different.  A negative is that
it accepts semantically \emph{different} programs that happen to behave the same
for certain inputs.  Furthermore, testing can be computationally expensive.
Automated testing requires a set of programs to use a
standardized build and test system for all their tests.
Flaky tests that fail non-deterministically can yield misleading results.
Most
importantly, testing requires a test suite with high coverage and strong
assertions that an LLM cannot
reward-hack.

% Behavioral equivalence through test execution offers practical semantic validation by comparing run-time behavior. This approach can detect subtle functional differences and provides confidence in correctness. However, this method is extremely repository-dependent, requiring specific build systems, test frameworks, dependency management, and environment configurations that vary dramatically across repositories and programming languages. Test-time comparison is fundamentally weaker than true semantic equivalence since it can only validate behavior for the specific test cases provided, potentially missing critical semantic changes that affect untested scenarios or edge cases. Models can learn to satisfy existing test suites while introducing subtle bugs or failing on inputs not covered by the tests, making this approach vulnerable to reward hacking where models optimize for test passage rather than true correctness. Additionally, it presents severe scalability challenges: requiring test infrastructure setup, managing diverse dependencies across repositories, executing potentially unsafe code, and consuming substantial computational time.

\smallskip

\label{sec:normalization}

Although semantic equivalence would be the best comparison method, it is
infeasible.
Therefore, we use source code equivalence as a
practical and scalable approximation that captures many cases of
functionally equivalent code.
Our normalization procedure removes
% For C-style languages (Java, JavaScript, TypeScript, C, C++, C\#, Rust,
% PHP), we remove both
block comments (e.g., \texttt{/* ... */}), line comments (e.g.,
\texttt{// ...}), and docstrings (e.g., \texttt{"""..."""}),
% and
% \texttt{\textquotesingle\textquotesingle\textquotesingle...\textquotesingle\textquotesingle\textquotesingle}),
using language-specific rules.
Our normalization procedure also standardizes whitespace.
It removes leading whitespace (except for whitespace-sensitive
languages like Python and Ruby) and trailing whitespace, and it collapses all
consecutive whitespace characters to a single space.

\section{LLMergeJ Training}

We trained LLMergeJ, a small but capable LLM for resolving Java
merge conflicts.
Our experiments (\cref{sec:results}) evaluated LLMergeJ against commercial LLMs, revealing the strengths and weaknesses of general-purpose commercial LLMs for resolving merge conflicts.

Due to resource constraints, we trained LLMergeJ focused on one programming language. The methodology readily scales to millions of samples given the vast repository of merge conflicts available in public repositories, because our approach requires no manual verification or annotation.

For training, we used the procedure of \cref{sec:dataset-construction} with the first 1000 starred Java
repositories, which are distinct from the Merge-Bench repositories.
This training set has \numJavaTrainRepositories Java repositories and
\numJavaTrainConflicts selected conflict hunks.

\textbf{Model Input.}
We used DeepSeek R1's system prompt and the query in
\cref{fig:merge-prompt}, adding the conflicting code and language marker.

\begin{figure}[t]
\centering
\fbox{%
  \begin{minipage}{\linewidth}
    \itshape\smaller
    You are a merge conflict resolution expert. Below is a snippet of code with surrounding context that includes a merge conflict.\\
    Return the entire snippet (including full context) in Markdown code syntax as provided.\\
    Do not modify the context at all and preserve the spacing as is.\\
    Think in terms of intent and semantics that both sides of the merge are trying to achieve.\\
    If you are not sure on how to resolve the conflict or if the intent is ambiguous, please return the same snippet with the conflict.\\
    Here is the code snippet: \\
    {\texttt{\textasciigrave\textasciigrave\textasciigrave<language>}} \\
    {\texttt{<conflict>}} \\
    {\texttt{\textasciigrave\textasciigrave\textasciigrave}}
  \end{minipage}%
}
\precaptionspace
\caption{User prompt for merge conflict resolution.}
\label{fig:merge-prompt}
\end{figure}

\textbf{Reward Structure.}
\label{sec:reward-structure}
LLMergeJ's reward function is the sum of three components
\(
R_{\text{total}} = R_{\text{reasoning}} + R_{\text{format}} + R_{\text{resolution}}
\)
where:

\vspace{-10pt}

\begin{align*}
R_{\text{reasoning}} &=
  1 \text{~if output uses \texttt{<think>} tokens},
  0 \text{~otherwise}
\\
R_{\text{format}} &=
1 \text{~if output has correct Markdown formatting},
0 \text{~otherwise}
\\
R_{\text{resolution}} &= \begin{cases}
1.0 & \text{if textual match with ground truth} \\
0.5 & \text{if source code match (normalized)} \\
0.1 & \text{if conflict preserved (no resolution)} \\
0 & \text{otherwise (likely incorrect merge)}
\end{cases}
\end{align*}

\begin{figure}[t]
\centering
\begin{minipage}[t]{0.48\columnwidth}
\centering
\includegraphics[width=\textwidth]{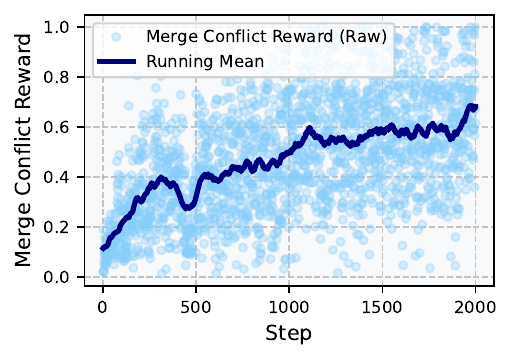}
\postgraphicsspace
\postgraphicsspace
\precaptionspace
\caption{Merge conflict reward over time during the training process.}
\label{fig:merge-conflict-reward}
\end{minipage}
\hfill
\begin{minipage}[t]{0.48\columnwidth}
\centering
\includegraphics[width=\textwidth]{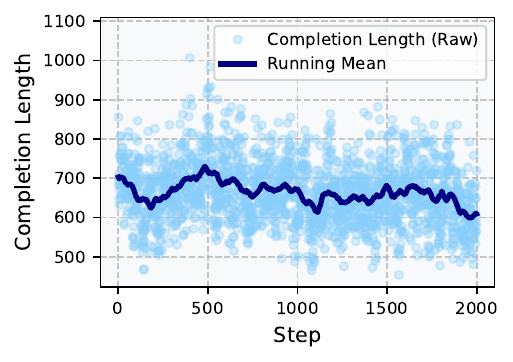}
\postgraphicsspace
\postgraphicsspace
\precaptionspace
\caption{The evolution of completion length over the training process.}
\label{fig:completion-length}
\end{minipage}
\end{figure}

\textbf{GRPO.} We trained our model using GRPO \cite{ShaoWZXSBZZLWG2025,DeepseekR1}, which is a simpler policy optimization approach than PPO~\cite{PPO} since it does not require the construction of a critic network. GRPO works by sampling a few rollouts from a network and reinforcing the higher-reward ones relative to the others.
To estimate the advantage, it uses the mean reward of all rollouts, which replaces the usual critic baseline with a group-normalized return.
For each prompt \(q\) we
\begin{flalign*}
\textrm{sample \(G\) outputs \quad}&
\{o_i\}_{i=1}^G \sim \pi_{\theta_{\rm old}}(\cdot\mid q)\ ,
&\\
\textrm{compute standardized advantages \quad}&
A_i \;=\;
\frac{r_i \;-\;\mathrm{mean}\bigl(\{r_1,r_2,\ldots,r_G\}\bigr)}
     {\mathrm{std}\bigl(\{r_1,r_2,\ldots,r_G\}\bigr)}\ ,
&\\
\textrm{and define the probability ratio \quad}&
\rho_i(\theta)
= \frac{\pi_\theta(o_i\mid q)}{\pi_{\theta_{\rm old}}(o_i\mid q)}\ .
\end{flalign*}

\noindent
The objective function of GRPO is then expressed as
\begin{align*}
J_{\rm GRPO}(\theta) &= \mathbb{E}_{q,\{o_i\}}\!\Biggl[
\frac{1}{G}\sum_{i=1}^G
\min\!\bigl(\rho_i(\theta)\,A_i,\, \mathrm{clip}(\rho_i(\theta),\,1-\epsilon,\,1+\epsilon)\,A_i\bigr) \\
&\quad -\;\beta\,D_{\mathrm{KL}}\bigl(\pi_\theta(\cdot\mid q)\,\parallel\,\pi_{\theta_{\rm old}}(\cdot\mid q)\bigr)
\Biggr]\ .
\end{align*}

\textbf{GRPO over Supervised Fine-Tuning.} To demonstrate the effectiveness
of our reinforcement learning approach, we compare GRPO against Supervised
Fine-Tuning (SFT\@).  SFT faces two fundamental challenges in merge
conflict resolution: the inability to optimize reasoning processes and the
unpredictability of ground truth solutions.

The most significant advantage of GRPO is its ability to fine-tune the
reasoning step. Our reward structure includes the traditional reasoning
reward, allowing GRPO to optimize not just the final merge resolution but
also the intermediate reasoning process. SFT, in contrast, can only learn
from the final output: it cannot improve the model's reasoning capabilities
during conflict analysis.\looseness=-1

Additionally, developers may resolve two similar conflict hunks in semantically equivalent but syntactically different ways.
For instance, a developer might choose to refactor the conflicting code, using a cleaner implementation that achieves the same functionality. SFT would create misleading gradients, potentially degrading the model's ability to generate alternative valid solutions.

GRPO addresses these challenges through its group-relative optimization
approach. When the ground truth is unpredictable due to multiple valid
resolutions, most model outputs will receive a reward of 0 (for incorrect
attempts), while outputs that preserve the conflict markers receive a small
positive reward of 0.1. The group averaging then creates gradients that
favor conflict preservation over incorrect resolution attempts.
It does not favor any one resolution over the others, while still allowing the model to learn from cases where the resolution is more predictable. Combined with the reasoning reward, this approach enables the model to develop better analytical capabilities while avoiding overfitting to specific syntactic patterns.

\textbf{Training Procedure.} We fine-tuned the DeepSeek-R1 14B distilled
Qwen model, which has been quantized by unsloth~\cite{unsloth} to \quantizationBits bits. This model
already has reasoning incorporated into it. To reduce training requirements
we use lora adapters with rank and alpha value of \loraRank; this allows us to
not train the entire network. We additionally limit the input prompt length
to \maxInputTokenLimit tokens and the output length to \maxOutputTokenLimit tokens. We set the model
temperature to \modelTemperature and the number of generations per sample to \numGenerations with a
batch size of \batchSize and gradient accumulation steps set to \gradAccumSteps. As optimizer we
use the paged adamw with $\beta_1=0.9$, $\beta_2=0.99$, weight decay 0.0, and learning rate 5e-5 with constant warmup over 30 steps. We use standard GRPO parameters: clipping $\epsilon=0.2$, KL coefficient $\beta=0.0$ (no reference model for memory efficiency), gradient clipping at max norm 0.2, and DAPO loss normalization. The training process took \trainingDays days on an H200
GPU with AMD EPYC 9534 64-Core Processor, 1.5TB memory, running Red Hat Enterprise Linux 9.5\@. \Cref{fig:merge-conflict-reward} shows how the model's merge conflict reward improves during training. The detailed plots of all reward components during training are provided in the technical appendix.

\Cref{fig:completion-length} shows how the completion length (which includes {\smaller\texttt{<think>}} tokens) changes during training. Notably, this graph remains relatively stable, unlike general-purpose models such as R1, where completion length increases during training~\cite{DeepseekR1}. This stability likely reflects our constrained experimental setup with limited token budgets and shorter training duration.

\section{LLM models evaluated}

\Cref{fig:model-performance} shows the LLMs that we evaluated on the task
of resolving Java conflicts, all prompted zero-shot with the query from
\cref{fig:merge-prompt} and no in-context examples.  Our procedure can be applied to any
programming language, but our small LLMergeJ 14B-parameter model has only been
trained on Java.  Java is most common in
prior research in merge conflict resolution~\cite{ApelLBLK2011,LessenichAL2014,LarsenFBM2023,ScheschFYRE2024}.

Our model has \modelParameters parameters.  We also trained a supervised
fine-tuning (SFT) baseline using Qwen3-14B~\cite{qwen3fewAuthors}.  We intentionally
use different base models for our comparison: LLMergeJ builds on
DeepSeek-R1 (a reasoning-capable model) while the SFT baseline uses
Qwen3-14B (a more recent and capable standard model). This design choice is
necessary because SFT cannot optimize reasoning processes (unless SFT is
run on reasoning traces including \codeid{<think>} tokens){\,---\,}it can
only learn from final outputs. Applying SFT to a reasoning model would fail
to leverage its reasoning capabilities, as SFT has no mechanism to improve
the intermediate reasoning steps that reasoning models generate. By using
the more recent Qwen3-14B for SFT, we ensure a fair comparison that, if
anything, favors the SFT baseline. This allows us to evaluate whether
GRPO's ability to optimize reasoning provides advantages over traditional
SFT even when SFT uses a stronger base model. The SFT baseline's training
set is the same as LLMergeJ and contains the ground truth developer
solutions from \numJavaTrainConflicts merge conflicts as its target.  We
conducted an extensive hyperparameter search
using Qwen3-14B as the base model, systematically
evaluating 24 different configurations across learning rates (1e-4, 1e-5, 1e-6), weight decay values (0, 0.01), learning rate schedulers (linear,
cosine), and training epochs (1, 3).  In \cref{fig:model-performance}, ``Qwen3-14B SFT'' represents the optimal configuration from this
comprehensive search, selected based on the highest percentage of correctly resolved merges on
the test set of \numJavaTestConflicts conflicts. The detailed results of each configuration are provided in the technical appendix. This process heavily favors SFT over GRPO\@. Distillation and SFT models also receive 4× higher token limits (8,192 vs.~2,048 output tokens).

\section{Results and Analysis}
\label{sec:results}

\begin{figure}[t]
\centering
\footnotesize
\begin{tabular}{l@{\hspace{8pt}}c@{\hspace{8pt}}c@{\hspace{8pt}}c@{\hspace{8pt}}c@{\hspace{8pt}}c}
\toprule
& \multicolumn{3}{c}{Comparison to developer}	&	& \\
\thead{Model}	& \thead{Equivalent\\text}	&\thead{Code normal-\\ized equivalent}	& \thead{Different\\code}	& \thead{Conflict (no\\resolution)}	& \thead{Invalid \\Markdown}\\
\midrule                                                                                                                                                                                                                              
Gemini 2.5 Pro	& \textbf{54.7\%}	& \textbf{62.5\%}	& 34.1\%	& \phantom{0}3.4\%	& \phantom{0}0.0\% \\
o3 Pro	& 46.1\%	& 54.3\%	& 35.1\%	& 10.6\%	& \phantom{0}0.0\% \\
Claude Opus 4	& 44.4\%	& 51.2\%	& 27.6\%	& 21.2\%	& \phantom{0}0.0\% \\
Grok 4	& 33.4\%	& 39.7\%	& 17.9\%	& 42.4\%	& \phantom{0}0.0\% \\
\midrule                                                                                                                                                                                                                              
Llama 4 Maverick	& 26.2\%	& 32.6\%	& 27.1\%	& 40.1\%	& \phantom{0}0.2\% \\
QwQ 32B	& 32.1\%	& 43.2\%	& 30.9\%	& 20.5\%	& \phantom{0}5.5\% \\
Qwen3 8B	& \phantom{0}5.5\%	& \phantom{0}9.1\%	& \phantom{0}4.7\%	& 86.1\%	& \phantom{0}0.1\% \\
Qwen3 14B	& 12.9\%	& 16.6\%	& \phantom{0}8.7\%	& 74.7\%	& \phantom{0}0.0\% \\
Qwen3 32B	& 13.2\%	& 16.9\%	& 11.1\%	& 71.8\%	& \phantom{0}0.1\% \\
Qwen3 235B	& 30.9\%	& 39.5\%	& 25.4\%	& 35.1\%	& \phantom{0}0.0\% \\
R1 1.5B	& \phantom{0}0.0\%	& \phantom{0}0.2\%	& 46.5\%	& 44.0\%	& \phantom{0}9.3\% \\
R1 8B	& \phantom{0}3.5\%	& \phantom{0}8.1\%	& 31.6\%	& 58.4\%	& \phantom{0}1.9\% \\
R1 14B	& \phantom{0}9.3\%	& 13.4\%	& 15.4\%	& 70.7\%	& \phantom{0}0.5\% \\
R1 32B	& 22.8\%	& 30.4\%	& 29.3\%	& 39.7\%	& \phantom{0}0.6\% \\
R1 70B	& 25.7\%	& 33.0\%	& 26.9\%	& 39.6\%	& \phantom{0}0.5\% \\
R1-0528 671B	& 35.9\%	& 42.4\%	& 24.0\%	& 33.2\%	& \phantom{0}0.4\% \\
Qwen3-14B SFT	& 36.6\%	& 44.7\%	& 36.1\%	& 19.2\%	& \phantom{0}0.0\% \\
LLMergeJ 14B	& \underline{48.8\%}	& \underline{58.9\%}	& 35.5\%	& \phantom{0}5.6\%	& \phantom{0}0.0\% \\
\bottomrule
\end{tabular}
\precaptionspace
\caption{Success in merging Java conflicts.
  ``Equivalent text'' is a subset of ``code normalized equivalent.
  The last 4 columns sum to 100\%.
  Best results are shown
  in \textbf{bold} (1st place) and \underline{underlined} (2nd place). Proprietary and public models are separated.}
\label{fig:model-performance}
\end{figure}

% LocalWords:  Qwen3 R1 nd

Our experimental infrastructure classifies a merge resolution into five categories:

\textbf{Equivalent Text:} The model's output exactly matches the ground truth resolution as a string, including all whitespace, comments, and formatting.

\textbf{Code Normalized Equivalent:} The model's output differs from the ground
truth only in formatting or comments. Note that Textual Match is a subset of Normalized Source Code Match.

\textbf{Conflict:} The model outputs the original conflict markers unchanged, indicating uncertainty about the correct resolution. This is considered a valid response when the merge intent is ambiguous, as instructed in our prompt.

\textbf{Different Code:} The model's normalized output differs from the normalized ground truth. This represents possibly incorrect resolution attempts.

\textbf{Invalid Markdown:} The model's output does not contain a Markdown
fenced code block (missing or malformed \texttt{```} markers).

The last four categories are mutually exclusive and collectively exhaustive.

\subsection{Java Experiments}

\begin{figure}[t]
\centering
\footnotesize
\setlength{\tabcolsep}{.5em} 
\begin{tabular}{lccccc}
\toprule
\thead{Model}	& \thead{Equivalent\\text}	&\thead{Code normal-\\ized equivalent}	& \thead{Different\\code}	& \thead{Conflict (no\\resolution)}	& \thead{Invalid \\Markdown}\\
\midrule
Gemini 2.5 Pro	& \textbf{47.1\%}	& \textbf{52.6\%}	& 42.1\%	& \phantom{0}5.3\%	& 0.0\% \\
o3 Pro	& 39.2\%	& \underline{45.1\%}	& 40.9\%	& 14.1\%	& 0.0\% \\
Claude Opus 4	& \underline{40.3\%}	& 44.8\%	& 34.5\%	& 20.4\%	& 0.3\% \\
Grok 4	& 27.7\%	& 31.7\%	& 20.9\%	& 47.3\%	& 0.1\% \\
Qwen3 235B	& 25.8\%	& 30.6\%	& 32.0\%	& 37.3\%	& 0.1\% \\
R1-0528 671B	& 32.0\%	& 36.5\%	& 26.3\%	& 36.9\%	& 0.4\% \\
\bottomrule
\end{tabular}
\precaptionspace
\caption{Model performance summary across all languages.
Columns are as in \cref{fig:model-performance}.}
\label{fig:model_performance_summary}
\end{figure}

\begin{figure*}[t]
\centering
\includegraphics[width=1.15\textwidth]{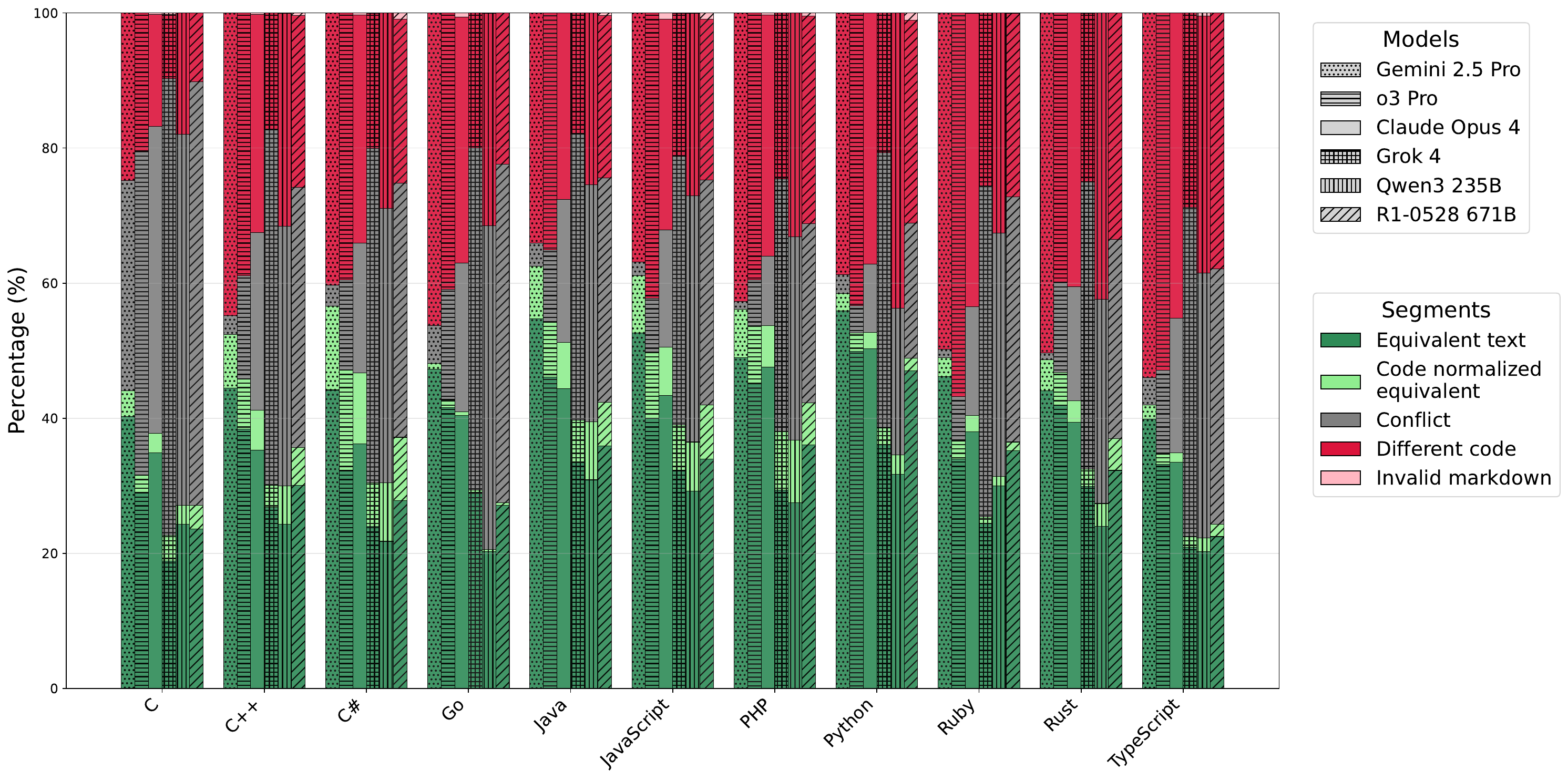}
\postgraphicsspace
\postgraphicsspace
\precaptionspace
\caption{Performance comparison across models and programming languages.}
\label{fig:multi-language}
\end{figure*}

Remarkably, our \modelParameters parameter LLMergeJ model is competitive with the
largest available models at the time of our experiments.  Our model ranks
second in textual and source code equality (\cref{fig:model-performance}).
Gemini 2.5 Pro is best.
LLMergeJ outperforms
commercial models like Claude Opus 4, o3 Pro, and Grok 4 that have orders of
magnitude more parameters. This represents a substantial
parameter efficiency gain.

The SFT baseline uses hyperparameters chosen to perform best on the
exact test data used for evaluation.  Despite this bias in its favor, the
SFT baseline outputs the same code as the developer only about
$\frac{3}{4}$ as often as our GRPO-trained model.
This demonstrates the benefit of reinforcement
learning for merge conflict resolution.

%% Data for "all of the models provide negative benefit":
% ;; Lists are: same, different, no_resolution
% (defvar qwen3-8b (list 9.1 4.7 86.2))
% (defvar llmergej (list 58.9 35.5 5.6))
% (defvar gemini25 (list 62.5 34.1 3.4))
% (defvar sft (list 44.7 36.1 19.2))
% (defun eval-with-k (k comparison)
%   (- (cl-first comparison) (* k (cl-second comparison))))
% ;; At k=2, they are all worse than nothing.  At k=1.5, three of the four
% ;; are better than nothing.  SFT is always worse than nothing.
% (let ((k 1.5))
%   (message " ")
%   ;; (message "qwen3-8b %s" (eval-with-k k qwen3-8b))
%   (message "sft %s" (eval-with-k k sft))
%   (message "llmergej %s" (eval-with-k k llmergej))
%   (message "gemini25 %s" (eval-with-k k gemini25))
% )

Correct merge resolutions are not the only important metric; a model should
be penalized for incorrect merge resolutions, which are worse for the
programmer than leaving the conflict in place~\cite{ScheschFYRE2024}.  If a
bad resolution is twice as bad as a correct resolution is good, then all of
the models provide negative benefit to a programmer.  It is straightforward
to make a model less aggressive (leaving the conflict in place more often)
by changing the training reward (currently 0.1 for leaving the conflict in
place, as noted in \cref{sec:reward-structure}) for LLMergeJ and SFT, or by
changing the prompts for the other models.

Our evaluation reveals three behavioral patterns among the
evaluated models, providing insights into different approaches to merge
conflict resolution.

\begin{itemize}
\item
  \textbf{Format-struggling models} (QwQ 32B, R1 1.5B) exhibit significant
  in\-struc\-tion-fol\-low\-ing deficits, producing \formatStrugglingInvalidMin--\formatStrugglingInvalidMax{}\% invalid outputs that fail to meet basic
  formatting requirements, indicating fundamental challenges in following
  the task specification. Surprisingly, this does not prevent the QwQ
  32B~\cite{qwq32b} model from being competitive.
\item
\textbf{Conservative resolvers} (such as Grok 4, Llama 4
Maverick, Qwen3 8B) preserve \conservativeResolversMin--\conservativeResolversMax{}\% of conflicts unresolved, suggesting
a cautious approach when facing ambiguous merge scenarios.
Outputting a conflict does no harm, but
their high conflict rates indicate
uncertainty in resolution capabilities.

\item
\textbf{Aggressive resolvers} (including our model LLMergeJ, Gemini 2.5 Pro, o3 Pro, and Claude Opus 4) demonstrate high confidence in their resolution capabilities, successfully resolving 51.2--62.5\% of conflicts with code normalized equivalent or better matches while preserving only 3.4--21.2\% of conflicts unresolved. These models exhibit excellent Markdown formatting following but very rarely maintain conflicts.

\end{itemize}

% We look forward to future models that outperform these models.

\subsection{Multi-language experiment}

The above experiment focused on resolving merge conflicts in Java programs.
Resource constraints prevented us from training LLMergeJ on
all languages.  Therefore, we do not know whether our results would
generalize to other programming languages.  To learn whether LLM-assisted
merging is different in other languages, we ran the best models against
the non-Java parts of Merge-Bench.

\Cref{fig:multi-language} shows that some languages, such as C, are harder
to merge; in fact, Java is one of the easiest languages to merge, along
with PHP and Python. Notably, for C language conflicts, we observe that models appear to be aware of the inherent challenges and adopt extremely careful merging strategies, preserving many conflicts rather than attempting potentially incorrect resolutions. More importantly, the general patterns of LLM
behavior, such as which LLMs are better than others and which LLMs give up
without resolving a conflict, are \emph{similar} across all programming
languages.  This is suggestive that our technique may generalize to other
languages.

\Cref{fig:model_performance_summary} summarizes the aggregate performance.
Gemini 2.5 Pro performs 7 percentage points better than the next-best
models, o3 Pro and Claude Opus 4.

The evaluation on commercial LLMs cost thousands of US dollars and weeks of time.  However,
running our small model LLMergeJ was fast and cheap.

\section{Limitations}

Like any research, ours has limitations.  The programmer merge might
have been wrong.  Although the prompt (a merge conflict) does not textually
appear on the Internet, we do not know how commercial training datasets are
created and they could conceivably include textual representations of merge
conflicts.  It is possible to choose only merge conflicts created after a
model's training cutoff date.  While the number of merges on GitHub is
vast, it is best to select from high-quality repositories (as Merge-Bench
does).  We have not directly evaluated our training approach on languages
other than Java.  The limitation in hunk size might exclude some of the
hardest merge problems, making our results an over-estimate of performance
on all merge problems.

\section{Conclusion}

This paper presents two major contributions to automated merge conflict
resolution: Merge-Bench and LLMergeJ.

\textbf{Merge-Bench} is a large-scale benchmark containing \numTotalConflicts
real-world merge conflicts from \numTotalRepositories repositories.  More
importantly, the Merge-Bench construction methodology is scalable to the
millions of repositories on GitHub.  Merge-Bench addresses fundamental
limitations in existing evaluation approaches.  It uses real-world data, it
minimizes data leakage, it is scalable (zero human labeling), and our
test-free evaluation paradigm avoids reward hacking vulnerabilities.

\textbf{LLMergeJ} is the first approach to leverage online reinforcement
learning for merge conflict resolution.  Evaluation on
Merge-Bench, demonstrated that our \modelParameters parameter model
performs competitively with much larger commercial models while
maintaining excellent parameter efficiency. Our methodology addresses
fundamental scalability limitations in existing approaches by eliminating
the need for human filtering or manual adjustments, enabling truly scalable
training on internet-scale data with no human annotation requirements.

\subsubsection{Acknowledgements}
This material is based upon work supported by the Defense Advanced Research
Projects Agency (DARPA) under Agreement No.\ HR00112590132.

\bibliographystyle{splncs04}
\bibliography{plume-bib/bibstring-abbrev,plume-bib/ernst,plume-bib/nlp,plume-bib/other,plume-bib/soft-eng,plume-bib/testing,plume-bib/version-control,plume-bib/crossrefs-abbrev}

\end{document}